\documentclass{article}
\usepackage{paperlayout,times}
\usepackage{amsmath,amssymb,amsfonts,bm,booktabs,graphicx,multirow,makecell,array,tabularx,longtable,xcolor,subcaption,enumitem,wrapfig,placeins,needspace,algorithm,algpseudocode,hyperref}
\hypersetup{hidelinks}
\usepackage{url}
\usepackage[nameinlink]{cleveref}
\makeatletter
\providecommand{\theHALG@line}{\thealgorithm.\arabic{ALG@line}}
\makeatother
\title{A GPU-Parallel Framework for Heterogeneous Multi-Task Reinforcement Learning}
\author{
  \textbf{Rui Zhang}\textsuperscript{1}\thanks{These authors contributed equally.} \quad
  \textbf{Qiwei Wu}\textsuperscript{2}\footnotemark[1] \quad
  \textbf{Zhengyu Zhang}\textsuperscript{1} \quad
  \textbf{Tao Li}\textsuperscript{1} \quad
  \textbf{Hongyu Zhou}\textsuperscript{1} \\
  \textbf{Xiang Li}\textsuperscript{3} \quad
  \textbf{Yunrong Guo}\textsuperscript{1} \quad
  \textbf{Junjie Lai}\textsuperscript{1} \quad
  \textbf{Renjing Xu}\textsuperscript{2}\thanks{Corresponding authors.} \quad
  \textbf{Weihua Zhang}\textsuperscript{1}\footnotemark[2] \\
  \normalfont
  \textsuperscript{1}NVIDIA (\url{https://nvidia.com}) \\
  \textsuperscript{2}The Hong Kong University of Science and Technology (Guangzhou) (\url{https://hkust-gz.edu.cn}) \\
  \textsuperscript{3}Tsinghua University (\url{https://mail.tsinghua.edu.cn})
}
\date{}
\newcommand{\method}{DGPO}
\newcommand{\benchmark}{Hebero}
\newcommand{\benchmarkfullname}{\textbf{He}terogeneous \textbf{Be}nchmark for \textbf{Ro}bot Learning}
\newcommand{\IWLower}{0.5}
\newcommand{\IWUpper}{2}
\newcommand{\IWSlope}{10}
\newcommand{\TaskEMAAlpha}{0.05}
\newcommand{\IWStartSteps}{1{,}000}
\newcommand{\BCFloor}{0.1}
\newcommand{\BCCeiling}{1}
\newcommand{\BCKappa}{2}
\newcommand{\BCBase}{1}
\newcommand{\BCHorizonSteps}{200{,}000}
\newcommand{\BCHorizonIterations}{12{,}500}

\newcommand{\PPORolloutSteps}{16}

\newcommand{\libero}{LIBERO}

\newcommand{\cmark}{\ensuremath{\checkmark}}\newcommand{\xmark}{\ensuremath{\times}}
\paperfinalcopy
\begin{document}\maketitle\lhead{}
\begin{abstract}
GPU-parallel simulation provides abundant robot interaction, but existing benchmarks rarely combine this scale with heterogeneous manipulation tasks and standardized multi-task RL evaluation.
We introduce \benchmark{} (Heterogeneous Benchmark for Robot Learning), a GPU-parallel Isaac Lab benchmark that enables efficient joint training and evaluation of a single policy across all 40 heterogeneous tasks.
Scaling experiments show that increasing parallel replicas per task improves success under a fixed wall-clock budget.
To support learning with sparse rewards and limited demonstrations, we propose Demonstration-Guided Policy Optimization (\method{}), which reuses demonstrations for dense tracking rewards and asymmetric value learning.
Its shared stack supports controlled comparisons of learner-specific demonstration interfaces within PPO.
Within DGPO framework, we introduce IW-ABC, which uses a lightweight per-task learning progress signal to coordinate adaptive behavior cloning (ABC), relaxing demonstration guidance with task progress, and importance weighting (IW), emphasizing lagging tasks in PPO updates.
With 50 demonstrations per task, IW-ABC achieves 90.1\% state-input mean success, outperforming the strongest baseline FAMO-ABC by 7.8 percentage points.
Its visual counterpart reaches 93.5\% mean success.
Real-world experiments further demonstrate that a single multi-task policy trained in simulation can successfully perform four tasks on a physical Piper robot.
The project page is available at \url{https://hebero-rl.github.io/}.
\end{abstract}
\section{Introduction}
\label{sec:introduction}

GPU-parallel reinforcement learning has made robot simulation a high-throughput training substrate, but much of this throughput is still spent on specialist policies: one objective, one policy, and one training run at a time.
The next scaling question is whether this infrastructure can support \emph{capability breadth}, where a single policy acquires many structured manipulation skills in one training process.
Multi-task reinforcement learning (MTRL) offers a natural path because shared task structure can improve representation learning, exploration, and data efficiency~\citep{deramo2020sharing, xu2024myopic}.
In massively parallel robot RL, however, sample collection is no longer the only bottleneck; value stability, task imbalance, and the use of prior data become central algorithmic challenges~\citep{tao2024maniskill3, joshi2025mtbench, janwani2026moplayground}.

Existing benchmarks provide complementary parts of this capability.
RLBench offers diverse tasks, demonstrations, and multi-task learning settings on CPU-based physics~\citep{james2020rlbench}; \libero{} adds structured suites for lifelong learning~\citep{liu2023libero}.
Later, MTBench brings Meta-World tasks to GPU-parallel state-based MTRL~\citep{yu2020metaworld,joshi2025mtbench}, while ManiSkill3 supports heterogeneous GPU simulation and visual RL with per-task reference learners~\citep{tao2024maniskill3}.
\textbf{\benchmark{}} (\benchmarkfullname{}) combines these capabilities in one Isaac Lab~\citep{mittal2025isaaclab} benchmark: multiple MTRL baselines jointly learn all 40 heterogeneous tasks through GPU-parallel interaction, with state and visual inputs, demonstration trajectories, and a standardized evaluation protocol (\cref{fig:main-overview}).
The project page is available at \url{https://hebero-rl.github.io/}.

Benchmark throughput alone does not solve sparse exploration or multi-task optimization.
Where Meta-World relies on manually engineered task rewards~\citep{yu2020metaworld}, we derive a reusable tracking reward from demonstrations.
For multi-task balancing, PCGrad manipulates per-task gradients~\citep{yu2020pcgrad}, while FAMO avoids this overhead through loss-based weighting~\citep{liu2023famo}.
We instead use a lightweight per-task learning progress signal, measured by per-task success-rate EMAs, to coordinate importance weighting (IW) toward lagging tasks and adaptive behavior cloning (ABC) that relaxes demonstration guidance with task progress and training time.
The \textbf{Demonstration-Guided Policy Optimization (\method{})} framework integrates these mechanisms with task-conditioned PPO~\citep{schulman2017ppo} and asymmetric value learning, using \textbf{IW-ABC} as its reference recipe.
Its shared infrastructure enables fair comparisons of demonstration-guided RL algorithms under a common PPO backbone, reward, observation, and evaluation protocol.
Reference learners vary how they use demonstrations, including BC initialization, DAPG-style regularization~\citep{rajeswaran2018dapg}, and an RFCL-inspired reset curriculum~\citep{tao2024rfcl}.

\begingroup
\raggedbottom
\setlength{\textfloatsep}{12pt plus 2pt minus 2pt}
\begin{figure}[t]
  \centering
  \includegraphics[width=0.8\linewidth]{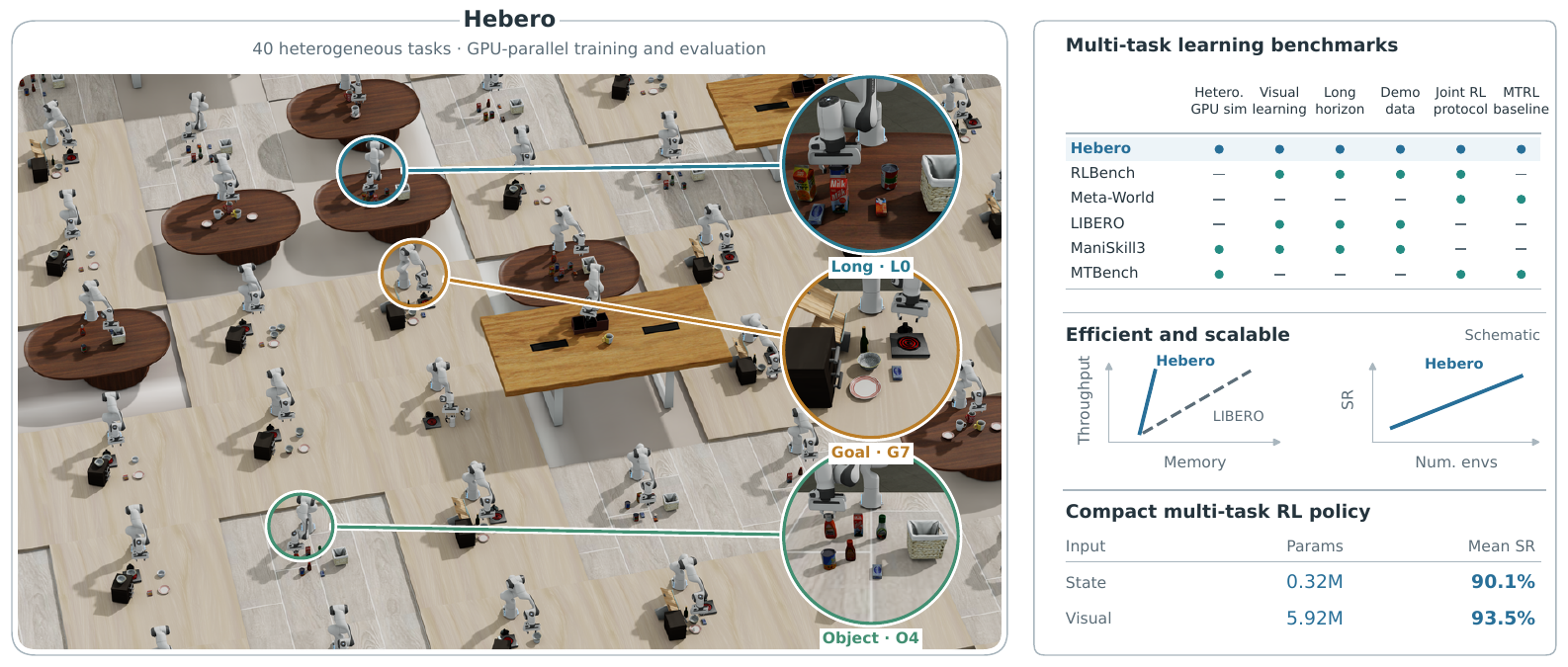}
  \caption{\textbf{\benchmark{}: heterogeneous tasks, scalable reinforcement learning, and compact policies.}
  \textbf{Left:} Forty \libero{} tasks share one GPU-parallel simulation, with magnified views highlighting distinct manipulation scenes.
  \textbf{Right:} Benchmark capabilities, schematic efficiency and scaling trends, and compact multi-task RL policies trained with the \method{} framework.}
  \label{fig:main-overview}
\end{figure}

Our experiments connect heterogeneous parallel simulation to joint learning with compact policies.
The schematic curves in \cref{fig:main-overview} illustrate memory-efficient throughput scaling and improved success with more parallel environments.
Across 40 tasks and 1,600 environments, \benchmark{} delivers approximately 7.5k simulation steps per second on one L20 GPU, 3.4 times the throughput of our 96-core MuJoCo baseline.
Increasing replicas per task improves success within a common wall-clock budget.
Within the shared \method{} stack, IW-ABC achieves the highest state-input Mean SR of 90.1\% with a 0.32M-parameter actor, outperforming the strongest baseline, FAMO-ABC (82.3\%), by 7.8 percentage points.
The visual policy reaches 93.5\% with 5.92M total deployed parameters, including its frozen encoder.
A separate jointly trained four-task policy transfers to a physical Piper robot with 82.5\% success and no weight updates.
Our contributions are threefold:
\begin{enumerate}[leftmargin=*, itemsep=2pt, topsep=2pt]
  \item \textbf{\benchmark{}: an efficient and scalable multi-task RL benchmark.} We compile all 40 \libero{} tasks into one heterogeneous GPU-parallel RL process with common state and visual interfaces, demonstrations, and evaluation protocols. GPU parallelism increases throughput, while scaling environments per task improves success under a fixed wall-clock budget. The project page is available at \url{https://hebero-rl.github.io/}.
  \item \textbf{Demonstration-guided training framework and reference recipe.} \method{} reuses demonstrations for dense tracking rewards and asymmetric value learning, providing a shared training stack with IW-ABC as its reference recipe. Applied to four Piper tasks, this recipe trains a joint state-based policy that transfers to the physical robot achieving 82.5\% success.
  \item \textbf{Adaptive imitation guidance and task balancing.} Within \method{}, IW-ABC uses the same lightweight per-task learning progress signal to guide both imitation and online RL across heterogeneous tasks. Compared with FAMO-ABC, the strongest state-input baseline, it improves Mean SR from 82.3\% to 90.1\% and Long SR from 53.3\% to 70.0\%.
\end{enumerate}
\endgroup
\section{Related Work}
\label{sec:related}

\subsection{GPU-Parallel Multi-Task Robot Learning}

Early CPU-based benchmarks established diverse manipulation tasks and learning protocols.
RLBench provided 100 tasks, including long-horizon compositions, with visual observations and motion-planned demonstrations for multi-task and few-shot learning~\citep{james2020rlbench}.
Contemporaneously, Meta-World defined joint multi-task and meta-RL evaluation over 50 tasks~\citep{yu2020metaworld}, later standardized by Meta-World+~\citep{metaworldplus2025}.
\libero{} organized demonstration-rich tasks into object, spatial, goal, and long-horizon suites for lifelong learning~\citep{liu2023libero}.
Their original execution models expose useful task diversity but do not provide a single GPU-parallel process for thousands of heterogeneous environments.
Subsequent GPU-accelerated systems scale simulation and learning: ManiSkill3 provides heterogeneous GPU physics and rendering~\citep{tao2024maniskill3}, RoboVerse unifies simulation and data interfaces~\citep{geng2025roboverse}, and MTBench ports Meta-World to Isaac Gym for state-based MTRL~\citep{joshi2025mtbench}.
Related directions include GPU-vectorized multi-objective RL in MO-Playground~\citep{janwani2026moplayground} and massively multitask RL on MMBench with the Newt world model~\citep{hansen2025newt}.
\benchmark{} brings 40 \libero{} tasks to a single heterogeneous GPU-parallel process for demonstration-guided joint RL, preserving their assets and success predicates while supporting state and visual policies under common training and evaluation interfaces.

\subsection{Demonstration-Guided and Multi-Task Reinforcement Learning}

Prior data can influence policy initialization, online optimization, the training-state distribution, or the reward signal.
DAPG uses demonstrations to initialize and regularize policy learning~\citep{rajeswaran2018dapg}, while RLPD incorporates offline data into online off-policy RL~\citep{ball2023rlpd} and Rainbow-DemoRL combines improvements to demonstration-augmented RL~\citep{bhatt2026rainbowdemorl}.
Data-Efficient Multitask DAgger allocates additional expert data according to task progress~\citep{fu2025multitaskdagger}.
RFCL instead uses demonstration states to construct a reverse-to-forward reset curriculum~\citep{tao2024rfcl}.
Reference tracking provides dense imitation-based rewards~\citep{2018-TOG-deepMimic}, while manipulation methods learn reusable dense rewards, combine demonstration-based reward and world-model learning, or estimate residual and interpretable rewards~\citep{mu2024drs,escoriza2025demo3,cao2025rrm,baimukashev2024transparent}.
Object-centric and reward-observation world models provide further training signals~\citep{ferraro2023focus,tang2025roboscaper}; adversarial imitation and multi-task inverse RL offer complementary ways to infer task objectives~\citep{kuang2025goalsagail,glazer2024commonsense}.
\method{} organizes these different uses of demonstrations through a shared trainability stack and additional learner-specific interfaces.

Multi-task optimization introduces a separate challenge: shared representations and task diversity can improve transfer and exploration~\citep{deramo2020sharing,xu2024myopic}, yet tasks may progress unevenly and produce conflicting updates.
Prior methods balance gradient magnitudes~\citep{chen2018gradnorm}, modify gradient geometry~\citep{yu2020pcgrad,liu2024cagrad}, or adapt task weights through loss dynamics~\citep{liu2023famo}.
PolicyGradEx estimates task affinities to group objectives for joint training~\citep{zhang2025policygradex}, while pessimistic offline RL addresses the risks of sharing data across task distributions~\citep{bai2024pessimistic}.
Our task importance weighting reweights PPO losses using lightweight online task-success statistics under fixed rollout allocation.
Its signal is independent of reward scale and separate from demonstration interfaces, allowing composition with the supported PPO references.
IW-ABC combines task prioritization with dynamic teacher guidance during online PPO training, adapting the strength of teacher supervision to each task's learning progress and overall training progress.
\section{The \benchmark{} Benchmark}
\label{sec:benchmark}

We separate the benchmark definition from the demonstration-guided reference learning framework.
\benchmark{} specifies a multi-task RL problem, a GPU-parallel environment construction, and a standardized interaction and evaluation contract.
\method{}, introduced in \cref{sec:method}, instantiates one demonstration-guided approach to this benchmark using the accompanying demonstration trajectories.

\subsection{Problem Setting}
\label{sec:problem-setting}

We follow the standard multi-task RL formulation~\citep{yu2020metaworld}, in which each task $T_k$ is a finite-horizon MDP $(\mathcal{S}_k,\mathcal{A},P_k,R_k,H,\gamma)$ drawn from a task distribution $p(T)$; notation is summarized in \cref{app:implementation}.
The tasks may differ in scene geometry, transition dynamics, reward, and success predicate, but share an action space, control horizon, and robot embodiment.
Multi-task RL learns one task-conditioned policy $\pi_\theta(a_t\mid o_t)$, where $o_t$ includes the task encoding $z_k$, to maximize average discounted return $\max_\theta\mathbb{E}_{T_k\sim p(T)}[\mathbb{E}_{\pi_\theta}[\sum_{t=0}^{H-1}\gamma^tR_k(s_t,a_t)]]$.
For \benchmark{}, $p(T)$ is uniform over $K=40$ tasks indexed by $k\in\{0,\ldots,K-1\}$.
We evaluate the single policy on these same tasks, rather than on a held-out task set.

\subsection{Benchmark Construction}
\label{sec:benchmark-construction}

\benchmark{} separates task semantics from simulator execution.
Offline task descriptors specify scene assets, reset rules, and success predicates.
Shared asset prototypes are instantiated in the environments that use them, with one batched view per prototype.
An environment-to-task map routes task-specific operations within one Isaac Lab simulator, renderer, rollout buffer, and policy update.
This heterogeneous GPU-parallel design and implementation enables efficient joint RL training across all 40 tasks, with throughput and wall-clock learning gains quantified in \cref{sec:benchmark-validation}.
Further construction details are provided in \cref{app:benchmark-construction}.

\subsection{Standardized Learning Interface}
\label{sec:benchmark-interface}

\benchmark{} exposes the same observation, action, reward, and success interfaces for every task, giving multi-task learners a common control contract across heterogeneous scenes.

\paragraph{Action space.}
At each step, the policy outputs a normalized action $a_t\in[-1,1]^7$.
The first six dimensions specify relative end-effector translation and rotation through an operational-space controller, and the last dimension controls the binary gripper.
Exact action scaling and controller parameters are given in \cref{app:action-controller}.

\paragraph{Actor observations.}
All actors receive proprioception $o_t^{\mathrm{proprio}}$, the previous action $a_{t-1}$, and a task encoding $z_k$.
State actors additionally receive a deployable object--target pose representation $o_t^{\mathrm{object\text{-}target}}$, whereas visual actors replace this representation with pooled frozen-ViT features $h_t^{\mathrm{visual}}$ from third-person and wrist RGB observations:
\begin{equation}
\begin{aligned}
  o_t^{\mathrm{state}}
  &=\left[a_{t-1},z_k,o_t^{\mathrm{proprio}},o_t^{\mathrm{object\text{-}target}}
           \right], \\
  o_t^{\mathrm{visual}}
  &=\left[a_{t-1},z_k,o_t^{\mathrm{proprio}},
           h_t^{\mathrm{visual}}\right].
\end{aligned}
\label{eq:actor-observations}
\end{equation}
Full observation dimensions, task encodings, and visual architecture are given in \cref{app:observations,app:visual-details}.

\paragraph{Task reward.}
Every task exposes learner-independent task reward $r_t^{\mathrm{task}}$ composed of three terms: 
a goal reward $r_t^{\mathrm{goal}}$ where $g_k(s_t)$ indicates task success, 
smoothness and control regularizers $r_t^{\mathrm{reg}}$ where $\dot q_t^{\mathrm{arm}}$ denotes arm-joint velocity,
and joint-limit penalties $r_t^{\mathrm{safe}}$:
\begin{equation}
\begin{aligned}
  r_t^{\mathrm{task}}
  &={}r_t^{\mathrm{goal}}-r_t^{\mathrm{reg}}-r_t^{\mathrm{safe}}, \\
  r_t^{\mathrm{goal}}
  &={}w_{\mathrm{succ}}\mathbf{1}[g_k(s_t)], \\
  r_t^{\mathrm{reg}}
  &={}w_{\Delta a}\lVert a_t-a_{t-1}\rVert^2
     +w_a\lVert a_t\rVert^2
     +w_{\dot q}\lVert\dot q_t^{\mathrm{arm}}\rVert^2, \\
  r_t^{\mathrm{safe}}
  &={}w_{\mathrm{pos}}^{\mathrm{lim}}\sum_j d_{j,t}^{\mathrm{pos}}
     +w_{\mathrm{vel}}^{\mathrm{lim}}\sum_j d_{j,t}^{\mathrm{vel}}.
\end{aligned}
\label{eq:task-reward}
\end{equation}
Detailed joint-limit penalties, and coefficient values are given in \cref{app:task-reward-details}.

\subsection{Evaluation Protocol}
\label{sec:benchmark-evaluation}
The benchmark evaluates placement robustness under a fixed, learner-independent reset distribution.
For each task and movable object, we compute coordinate-wise bounds $[x_{\min},x_{\max}]\times[y_{\min},y_{\max}]$ from the 50 recorded demonstration starts.
Each evaluation episode independently redraws the object's $x$ and $y$ coordinates uniformly within this box rather than replaying a recorded placement.
Collision-based rejection sampling is applied to the sampled initial placements to ensure that objects do not collide with each other.
Compared with \libero{}'s fixed evaluation starts~\citep{liu2023libero}, resampling better tests placement robustness rather than trajectory memorization.
We report \textbf{mean success rate (Mean SR)} across all tasks and suite-level SR, using \textbf{Long SR} as the long-horizon measure.
\textbf{Task coverage} is the number of tasks reaching at least 80\% success and measures the breadth of reliably solved skills.
\textbf{SR-AUC} is the normalized trapezoidal area under checkpoint-evaluated Mean SR as a function of PPO iteration over the shared training horizon, reported as a percentage of the maximum possible area.
It reflects learning speed and overall success.

\section{DGPO: A Demonstration-Guided Multi-Task Training Framework}
\label{sec:method}

GPU-parallel simulation accelerates interaction, but learning across \benchmark{}'s 40 heterogeneous tasks remains challenging under sparse goal rewards.
We therefore introduce \method{}, a training framework that leverages the available demonstrations to make joint learning practical through shared reward, reset, and critic guidance coupled to task-conditioned PPO.

\subsection{Shared Training Components}
\label{sec:shared-protocol}
All online reference configurations use the same demonstration command stream, tracking reward, privileged critic inputs, task-conditioned PPO backbone and network capacity.
Together these components form the shared DGPO trainability stack which supports controlled comparisons of learner-specific demonstration interfaces under matched training conditions.

\paragraph{Demonstration-trajectory command stream.}
The command stream centralizes demonstration bookkeeping across heterogeneous parallel environments.
It maintains each environment's task identity, active demonstration, and timestep cursor, and caches the corresponding robot, object, gripper, and action references from a shared trajectory bank.
These synchronized views let the tracking reward, critic, and learner interfaces reuse the same references without duplicating trajectory storage or cursor management.

\paragraph{Demonstration-tracking reward.}
Reference-based manipulation requires preserving robot--object and robot--environment interactions, as emphasized by OmniRetarget~\citep{yang2026omniretarget}.
We therefore track both robot motion and task-relevant world states, including rigid-object poses and articulation joint positions.
Let $\mathcal{E}_k^{\mathrm{robot}}$ contain task-$k$ end-effector pose, arm-joint position and velocity, and gripper-state errors; let $\mathcal{E}_k^{\mathrm{world}}$ contain rigid-object pose and articulation joint-position errors; and define $\mathcal{E}_k=\mathcal{E}_k^{\mathrm{robot}}\cup\mathcal{E}_k^{\mathrm{world}}$.
For component error $e_t^m$, weight $w_m$, and component-specific bandwidth $\sigma_{r,m}>0$, the demonstration-tracking reward is
\begin{equation}
  r_t^{\mathrm{demo}}
  =\sum_{m\in\mathcal{E}_k}w_m
    \exp\!\left(-\frac{\lVert e_t^m\rVert}{\sigma_{r,m}}\right),
  \label{eq:demo-tracking-reward}
\end{equation}
and the reference online configuration adds this term to the task reward in \cref{eq:task-reward}.
The dense terms reward both how the robot moves and what its motion accomplishes in the world.
Tracking weights $w_m$ and component-specific kernel bandwidths are listed in \cref{tab:app-parameters}.

\paragraph{Asymmetric critic observations.}
The state critic augments the state actor input as $o_t^V=[o_t^{\mathrm{state}},e_t^{\mathrm{robot}},e_t^{\mathrm{world}},\phi_t]$.
Here $e_t^{\mathrm{robot}}$ contains privileged end-effector, arm-joint position, and gripper tracking errors; $e_t^{\mathrm{world}}$ contains object-pose and articulation-position tracking errors;
and $\phi_t$ is the normalized current demonstration cursor (appendix \cref{eq:normalized-demo-cursor}).
The visual critic input is $o_t^V=[a_{t-1},z_k,o_t^{\mathrm{proprio}},h_t^{V,\mathrm{visual}},e_t^{\mathrm{robot}},e_t^{\mathrm{world}},\phi_t]$, where $h_t^{V,\mathrm{visual}}$ uses a separate trainable attention pool over the same frozen image tokens (\cref{app:visual-details}).

\subsection{Reference Learners and Demonstration Interfaces}
\label{sec:demonstration-interfaces}

\begin{wraptable}[8]{R}{0.49\textwidth}
  \vspace{-0.8\baselineskip}
  \centering
  \normalsize
  \setlength{\tabcolsep}{3pt}
  \renewcommand{\arraystretch}{0.92}
  \begin{tabularx}{\linewidth}{@{}lX@{}}
    \toprule
    \textbf{Learner family} & \textbf{Demonstration Entry} \\
    \midrule
    PPO & none \\
    BC$\rightarrow$PPO & offline actor initialization \\
    DAPG & offline likelihood regularizer \\
    ABC & live time-aligned action target \\
    RFCL & demo-state reset curriculum \\
    \bottomrule
  \end{tabularx}
\end{wraptable}
DGPO holds the shared components in \cref{sec:shared-protocol} fixed while varying the demonstration interface shown in the map; PPO adds no further demonstration mechanism.
Our RFCL entry applies an RFCL-inspired reset curriculum to on-policy PPO.
Reference-learner objectives, reset procedures, and settings are given in \cref{app:reference-learners}.
We next specify IW-ABC for the state and visual configurations.

\begin{figure}[tb]
  \centering
  \includegraphics[width=0.9\linewidth]{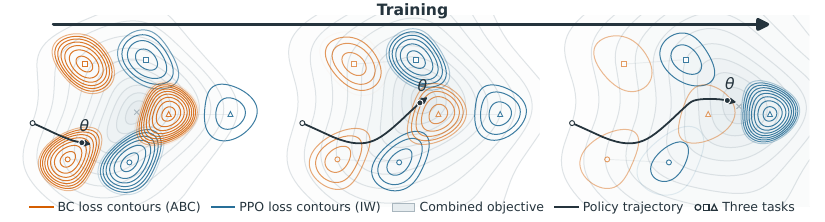}
  \caption{\textbf{Adaptive guidance and task balancing.}
  In a schematic parameter space, orange and blue contours show weighted per-task BC and PPO losses; matching marker shapes identify the same task.
  From early to late training, ABC guidance recedes at task-dependent rates while IW redistributes PPO emphasis, changing the combined objective (gray).
  The dark path illustrates how gradient updates move the shared policy $\theta$ as the combined objective evolves.}
  \label{fig:guidance-landscape}
\end{figure}

\subsection{IW-ABC Reference Recipe}
\label{sec:reference-recipe}

Demonstration-guided multi-task learning faces two coupled challenges: BC and PPO can favor competing policy updates, while tasks sharing one policy learn at different rates.
IW-ABC uses a single lightweight learning progress signal to regulate both imitation pressure and task priority: ABC relaxes action guidance as task success and training time increase, while IW gives greater PPO weight to tasks lagging behind.
\Cref{fig:guidance-landscape} illustrates how these changing weights reshape the combined objective for a shared policy, motivating the ABC and IW mechanisms below.

\paragraph{Task success as a progress signal.}
For each task $k$, the first batch of completed training episodes initializes $\tau_k\leftarrow\overline{\mathrm{SR}}_k$; subsequent batches update
\begin{equation}
  \tau_k \leftarrow
  (1-\alpha_{\mathrm{ema}})\tau_k
  +\alpha_{\mathrm{ema}}\,\overline{\mathrm{SR}}_k,
  \label{eq:success-ema}
\end{equation}
Here $\tau_k\in[0,1]$ is task $k$'s learning progress with update rate $\alpha_{\mathrm{ema}}$, 
and $\overline{\mathrm{SR}}_k$ is the mean binary success of the newly completed episodes.
Computed directly from training rollouts, this scalar provides a lightweight proxy for per-task learning progress without additional evaluation (\cref{app:dgpo-training}).

\paragraph{Adaptive action guidance.}
Let $u$ count vectorized rollout steps, advancing once per control step across all parallel environments, and let $U_{\mathrm{ann}}$ be the annealing horizon.
With annealing fraction $\rho_u=\min(u/U_{\mathrm{ann}},1)$, the coefficient is
\begin{equation}
  \beta_{k,u}
  =\beta_{\min}+(\beta_{\max}-\beta_{\min})
    \big[(1-\tau_k)(1-\rho_u)\big]^{\kappa}.
  \label{eq:beta-base}
\end{equation}
An early-solved task can therefore relinquish imitation before the global schedule ends, while time annealing prevents a persistently difficult task from remaining indefinitely at maximum guidance.
For rollout samples $(o_t,a_t^\star)$ with task index $k$, ABC adds
\begin{equation}
\mathcal{L}_{\mathrm{ABC}}
  =c_{\mathrm{BC}}\mathbb{E}_{t,k}\!\left[
    \beta_{k,u}\|\mu_\theta(o_t)-a_t^\star\|^2
  \right],
\label{eq:abc}
\end{equation}
where $\mu_\theta$ is the Gaussian policy mean and $a_t^\star$ is the demonstration action at the environment's current trajectory cursor.
The exact target construction is given in \cref{app:matched-action}.

\paragraph{Task balancing.}
Balanced collection does not ensure balanced learning: easy and persistently difficult tasks otherwise receive equal loss weight.
For initialized tasks $\mathcal{K}_{\mathrm{init}}$, we define the mean success $\bar{\tau}$, a sigmoid-transformed relative-success score $p_k^{\mathrm{IW}}$, and the importance weight $w_k$:
\begin{equation}
  \bar{\tau}=\frac{1}{|\mathcal{K}_{\mathrm{init}}|}
  \sum_{j\in\mathcal{K}_{\mathrm{init}}}\tau_j,
  \qquad
  p_k^{\mathrm{IW}}=\sigma\!\left(s_{\mathrm{IW}}(\tau_k-\bar{\tau})\right),
  \qquad
  w_k=w_{\max}(1-p_k^{\mathrm{IW}})+w_{\min}p_k^{\mathrm{IW}}.
  \label{eq:iw}
\end{equation}
where $\sigma$ is the logistic sigmoid and $s_{\mathrm{IW}}>0$ controls the sensitivity to relative task success.
Tasks below the multi-task mean receive larger weights.
IW-ABC minimizes the combined objective using minibatch-normalized weights $\tilde w_k$ (\cref{app:dgpo-training}):
\begin{equation}
  \mathcal{L}^{\mathrm{IW\text{-}ABC}}
  =\mathbb{E}_{t,k}\!\left[
    \tilde w_k\left(
      \ell_t^{\pi}+c_V\ell_t^V-c_H\mathcal{H}_t
    \right)
  \right]+\mathcal{L}_{\mathrm{ABC}},
  \label{eq:iw-abc}
\end{equation}
where $\ell_t^{\pi}$ is the negative clipped PPO surrogate, $\ell_t^V$ is the clipped value-regression loss, and $\mathcal{H}_t$ is policy entropy. 
$c_V$ and $c_H$ are the value and entropy coefficients.
Task IW scales all three PPO terms (ablation in appendix \cref{tab:app-iw-placement}).
Detailed hyperparameters are given in \cref{tab:app-parameters}.

FAMO~\citep{liu2023famo} offers another lightweight approach to task balancing, adapting task weights from loss progress.
Following FAMO's original actor--critic setting, we apply its task weighting only to critic losses and retain ABC, yielding FAMO-ABC (\cref{app:famo-abc}).
We compare IW-ABC and FAMO-ABC under the same training budget in \cref{sec:main-results}.

\begingroup\setlength{\textfloatsep}{12pt plus 2pt minus 2pt}\setlength{\intextsep}{8pt plus 2pt minus 2pt}\captionsetup{skip=4pt}
\section{Experiments}
\label{sec:experiments}

\subsection{Scaling Efficiency of the Hebero RL Benchmark}
\label{sec:benchmark-validation}

\paragraph{High throughput with a practical memory footprint.}
\Cref{tab:sim-throughput} shows that \benchmark{} supports higher simulation throughput than the evaluated MuJoCo configuration with a practical GPU memory footprint, and scales to multi-GPU end-to-end training across all 40 tasks.

\begin{table}[!t]
  \centering
  \captionsetup{skip=4pt}
  \caption{\textbf{Multi-task parallel simulation and RL training throughput (SPS).} Approximate MTBench results are from \citet{joshi2025mtbench}; hardware and peak memory were not reported.}
  \label{tab:sim-throughput}
  \normalsize
  \setlength{\tabcolsep}{2.5pt}
  \begin{tabular}{@{}l l r r r r@{}}
    \toprule
    \textbf{Benchmark} & \textbf{Engine (hardware)} & \textbf{Tasks} & \textbf{Envs} & \textbf{Throughput} & \textbf{Memory} \\
    \midrule
    \multicolumn{6}{l}{\emph{Simulation / collection}} \\
    \benchmark{} All-40 & Isaac Lab (1$\times$L20) & 40 & 1{,}600 & 7{,}500 & 10.6\,GiB VRAM \\
    \libero{} All-40 & MuJoCo (96C/192T CPU) & 40 & 1{,}600 & 2{,}200 & 1.5\,TiB host PSS \\
    \midrule
    \multicolumn{6}{l}{\emph{End-to-end RL training}} \\
    \benchmark{} All-40 & Isaac Lab (8$\times$L20) & 40 & 25{,}600 & 78{,}500 & 144.9\,GiB VRAM \\
    MTBench MT50-rand & Isaac Gym (N/A) & 50 & 24{,}576 & 69{,}000 & N/A \\
    \bottomrule
  \end{tabular}
\end{table}

\paragraph{Parallel scaling improves joint learning efficiency.}
Scaling the number of parallel environments translates simulation throughput into better joint learning: on a single 48\,GB L20 GPU, more replicas per task yield higher success within the same wall-clock budget (\cref{fig:parallel-env-scaling}(a)) and reach matched reward levels sooner (\cref{fig:parallel-env-scaling}(b)).
Thus, \benchmark{}'s parallel design enables faster learning across all 40 heterogeneous tasks.

\par\smallskip
\noindent\begin{minipage}{\linewidth}
  \centering
  \includegraphics[width=0.75\linewidth]{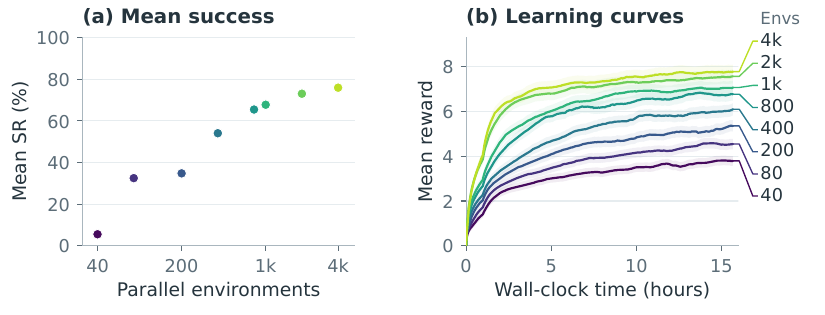}
  \captionof{figure}{\textbf{Single-GPU parallel environment scaling under a fixed wall-clock budget.}
  (a) Mean SR across all 40 tasks, evaluated at the common training-budget endpoint as parallel environments increase.
  (b) Mean training reward versus wall-clock training time for each environment count.}
  \label{fig:parallel-env-scaling}
\end{minipage}
\par\smallskip

\FloatBarrier
\Needspace{4\baselineskip}
\subsection{Controlled Learner Comparison within DGPO}
\label{sec:main-results}
\label{sec:setup}

\begin{wrapfigure}[19]{r}{0.45\linewidth}
  \vspace{-1.0\baselineskip}
  \centering
  \captionsetup{skip=4pt}
  \includegraphics[width=\linewidth]{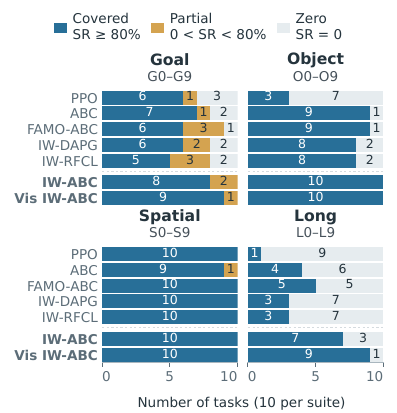}
  \caption{\textbf{Task coverage by suite.}
  Each bar partitions ten tasks by final SR.}
  \label{fig:task-coverage}
  \label{fig:task-aligned-success}
\end{wrapfigure}

All learners in \Cref{tab:main-results} jointly train on 40 tasks for 30{,}000 online PPO iterations with the same 50 demonstrations per task and undergo unified evaluation protocol (\cref{sec:benchmark-evaluation}).
State-input learners share a compact task-conditioned $[512,256,128]$ actor--critic.
We train each configuration with three training seeds; each run uses eight NVIDIA L20 GPUs with 2{,}000 parallel environments per GPU and takes approximately two days.
Detailed reproducibility settings are provided in \cref{tab:app-parameters} in the appendix.

\paragraph{Reference learner comparison.}
ABC improves all metrics over BC$\rightarrow$PPO, highlighting the value of online adaptive demonstration guidance beyond offline BC initialization.
With task IW fixed, IW-ABC outperforms IW-DAPG and IW-RFCL across all four metrics, supporting adaptive matched-action guidance over offline demonstration regularization and reverse-reset curricula for demonstration-guided joint learning.
Adding IW to PPO trades lower Mean SR for higher Long SR, whereas pairing it with ABC improves both.
With comparable computational complexity and training throughput, IW-ABC outperforms FAMO-ABC under the same training budget, showing the benefit of success-based IW as a simple, computationally efficient task-balancing mechanism when paired with ABC.
\Cref{fig:task-coverage} shows how these gains extend across task suites rather than only raising average success.
ABC, IW-ABC, and Vis IW-ABC cover 4, 7, and 9 of the ten Long tasks, demonstrating broader coverage of long-horizon skills.

\par
\ifnum\value{WF@wrappedlines}>3
  \vspace{\dimexpr\value{WF@wrappedlines}\baselineskip-3\baselineskip\relax}
\fi
\WFclear

\begin{table}[!hbp]
  \centering
  \captionsetup{width=0.95\linewidth}
  \caption{\textbf{PPO-based reference learner results.}
  All rows share DGPO framework; demonstration interfaces and task weighting vary.}
  \label{tab:main-results}
  \normalsize
  \setlength{\tabcolsep}{7pt}
  \begin{tabular}{@{}lccccc@{}}
    \toprule
    \textbf{Learner} & \textbf{IW} & \textbf{Mean SR (\%)} & \textbf{Long SR (\%)} & \textbf{Task coverage} & \textbf{SR-AUC (\%)} \\
    \midrule
    PPO & \xmark & 50.8$\pm$3.1 & 10.0$\pm$0.0 & 20/40 & 45.2 \\
    BC$\rightarrow$PPO & \xmark & 47.5$\pm$2.6 & 20.0$\pm$0.0 & 18/40 & 40.8 \\
    ABC & \xmark & 75.5$\pm$3.6 & 40.0$\pm$0.0 & 29/40 & 71.9 \\
    FAMO-ABC & \xmark & 82.3$\pm$2.9 & 53.3$\pm$5.8 & 30/40 & 75.6 \\
    \midrule
    IW-PPO & \cmark & 44.9$\pm$2.5 & 20.0$\pm$0.0 & 18/40 & 41.3 \\
    IW-DAPG & \cmark & 68.5$\pm$3.0 & 30.0$\pm$0.0 & 27/40 & 62.4 \\
    IW-RFCL & \cmark & 66.4$\pm$2.8 & 30.0$\pm$0.0 & 26/40 & 60.1 \\
    \textbf{IW-ABC} & \cmark & \textbf{90.1$\pm$3.8} & \textbf{70.0$\pm$10.0} & \textbf{35/40} & \textbf{82.1} \\
    \midrule
    \textbf{Vis IW-ABC} & \cmark & \textbf{93.5$\pm$2.6} & \textbf{81.9$\pm$11.5} & \textbf{38/40} & \textbf{83.3} \\
    \bottomrule
  \end{tabular}
\end{table}

\subsection{Learning Dynamics of IW-ABC}
\label{sec:weight-dynamics}

We examine the training dynamics of Vis IW-ABC to assess whether demonstration guidance recedes with task progress while PPO updates remain focused on lagging tasks; \cref{fig:weight-dynamics} traces task success alongside IW and BC coefficients.

\begin{figure}[!htbp]
  \centering
  \includegraphics[width=\linewidth]{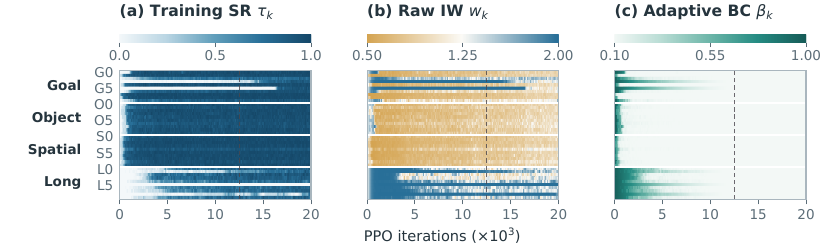}
  \caption{\textbf{Task-wise weighting dynamics of Vis IW-ABC.}
  Training SR EMAs (a) and reconstructed raw IW (b) and BC (c) coefficients from a single training run, averaged in 100-iteration bins.
  Rows retain task IDs 0--9 within each suite.
  Dashed lines mark the end of ABC annealing.}
  \label{fig:weight-dynamics}
\end{figure}

\paragraph{Prioritizing lagging tasks.}
IW reweights PPO losses under fixed rollout counts and optimizer steps.
Most Object and Spatial tasks succeed early, whereas the harder Goal and Long tasks learn more slowly and require greater optimization emphasis.
IW assigns these lagging tasks larger weights, amplifying their contribution to the shared PPO gradient until their success approaches the multi-task mean.

\paragraph{Annealing BC while retaining task priority.}
Once $u\geq U_{\mathrm{ann}}$, all BC coefficients reach the residual floor $\beta_{\min}$, relaxing constraints on deviation from demonstration actions while IW continues prioritizing lagging tasks (Fig.~\ref{fig:weight-dynamics}; see also Appendix Fig.~\ref{fig:app-weight-trajectories}).
Together, ABC regulates how strongly each task follows demonstrations, while IW directs PPO updates toward tasks that still need improvement, coupling a gradual release from imitation with continued support for harder tasks.
\subsection{Multi-Task Deployment on a Physical Robot}
\label{sec:real-world}

To evaluate IW-ABC's real-world deployability within \method{}, we jointly train one policy on four representative RoboTwin Piper tasks~\citep{mu2025robotwin}, selected for diverse interactions beyond repeated pick-and-place variants (\cref{fig:real-qualitative}).
Training uses parallel Isaac Lab replicas and 50 demonstration trajectories per task.
Since our focus is not visual-policy sim-to-real transfer, we deploy the state-input policy on the physical Piper robot using an external model-based perception pipeline (\cref{app:real-world}), with policy weights fixed throughout deployment.
Across 20 physical trials per task, the policy achieves an overall success rate of 82.5\%.
\Cref{fig:real-qualitative} shows representative execution sequences and per-task success counts.
Joint simulation training thus yields a deployable multi-task policy.

\begin{figure}[ht]
  \centering
  \includegraphics[width=0.97\linewidth]{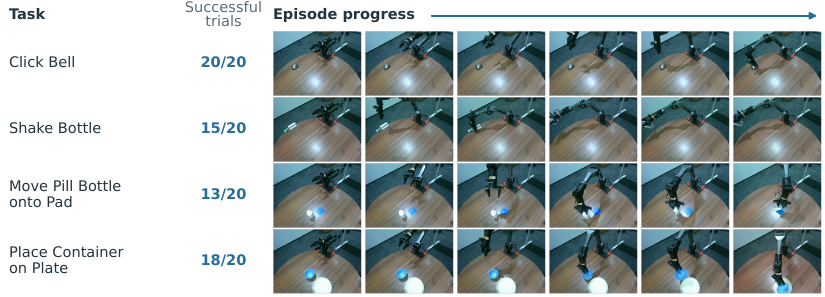}
  \caption{\textbf{Real-world deployment of one jointly trained four-task policy.}
  The simulation-trained state-based policy is deployed without weight updates on Piper robot.
  Rows show the four tasks with success counts over 20 trials each; columns show episode progress. Overall success is 82.5\%.}
  \label{fig:real-qualitative}
\end{figure}
\FloatBarrier
\endgroup
\section{Conclusion and Limitations}
\label{sec:conclusion}
\label{sec:limitations}

\benchmark{} supports joint GPU-parallel RL on all 40 \libero{} tasks; increasing replicas per task improves learning within a fixed wall-clock budget.
\method{} provides a shared demonstration-guided training stack for comparing PPO-based learners.
IW-ABC adapts teacher guidance and task weights using a cheap learning progress signal, improving mean and long-horizon success.
One jointly trained state-input policy achieves 82.5\% success on four physical Piper tasks without weight updates.
However, IW-ABC's time-indexed action targets can misalign under trajectory drift despite the tracking reward, and budget constraints limit training and evaluation to \libero{} and RoboTwin.
Future work will extend the benchmark construction methodology and DGPO training framework to more challenging tasks and larger task sets.
\clearpage\bibliography{main_arxiv}\bibliographystyle{paperrefs}\clearpage\appendix
\section{\benchmark{} Benchmark Specification and Implementation}
\label{app:implementation}

This section specifies the benchmark artifact independently of any particular learner: task compilation, heterogeneous simulation, controller, observation, reward, success, and randomized evaluation interfaces.
The accompanying DGPO framework and reference learners are documented separately in \cref{app:dgpo-reference}.

\paragraph{Problem notation.}
In \cref{sec:problem-setting}, task $T_k$ has state space $\mathcal{S}_k$, shared action space $\mathcal{A}$, transition kernel $P_k$, reward function $R_k$, episode horizon $H$ in control steps, and discount factor $\gamma$.
The distribution $p(T)$ samples tasks, indexed by $k$, from the $K$-task set.
The policy parameters are $\theta$; $s_t$, $o_t$, and $a_t$ denote simulator state, actor observation, and action at control step $t$, while $z_k$ is the task encoding.
The inner expectation in the return objective averages over policy-induced trajectories for the sampled task.

\subsection{Task Compilation and Heterogeneous Simulation}
\label{app:benchmark-construction}

\paragraph{Offline task descriptors.}
Before training, each \libero{} BDDL task is compiled into a cached JSON descriptor containing its instruction, scene, assets, spawn regions, goal predicates, and thresholds.
The descriptor governs task encoding, observations, resets, success checks, demonstration lookup, and logging; the original BDDL path is retained for provenance.

\paragraph{Goal predicates.}
Relational predicates (e.g., \texttt{in}, \texttt{on}), articulation predicates (e.g., \texttt{open}, \texttt{close}), and activation predicates (e.g., \texttt{turn-on}) are evaluated from live simulator state.
Task success is their conjunction and supplies the common sparse-reward event and evaluation outcome.

\paragraph{Asset pipeline.}
MuJoCo MJCF assets are converted offline to cached USD files while preserving geometry, appearance, dynamics, joints, and articulation limits.
Runtime descriptors resolve objects and fixtures to shared assets while retaining task-specific placement and randomization.

\paragraph{Heterogeneous vectorized simulation.}
Each environment instantiates the assets specified by its assigned task, while all environments share one simulator and renderer and expose a rectangular interaction batch to a single rollout and update pipeline.
For $K$ tasks and $g$ environments per task, the total environment count is $N=Kg$; the current implementation assigns environment $i\in\{0,\ldots,N-1\}$ to task $i\bmod K$.
This environment-to-task map routes reset rules, demonstration trajectories, goal predicates, and statistics for both state and visual policies.

\paragraph{Shared asset views.}
Each distinct asset prototype has one batched simulator view spanning exactly the environments in which it is instantiated.
Prototypes are shared across tasks and suites by asset model, scale, and physical configuration, with separate instance slots when a scene contains multiple copies of the same model.
Static fixtures additionally distinguish their fixed poses.
A view can cover a noncontiguous subset of the full environment batch; cached mappings between global environment IDs and local view indices support state gathering and reset writes.
Task-specific bindings retain each object's initial pose and semantic role, while observations are assembled into the common environment batch.

\paragraph{Task-set configuration.}
The canonical benchmark contains ten tasks from each of Goal, Object, Spatial, and Long.
The figure labels G$i$, O$i$, S$i$, and L$i$ preserve the logged within-suite index $i\in\{0,\ldots,9\}$ for \texttt{libero\_goal}, \texttt{libero\_object}, \texttt{libero\_spatial}, and \texttt{libero\_10}, respectively.
The active task set can be restricted for focused experiments without changing the canonical task registry.
Focused encoding restricts the task encoding and object--target pose buffer to the active subset, reducing the input dimension for single-suite and ablation runs.
Canonical encoding preserves the full benchmark dimensionality, allowing checkpoints to be loaded across task subsets without an input-dimension mismatch.

\subsection{Action and Controller Interface}
\label{app:action-controller}

The normalized action $a_t\in[-1,1]^7$ commands relative end-effector motion through an operational-space controller:
\begin{equation}
  \Delta p_t = s_p a_{t,1:3},
  \qquad
  \Delta R_t = \operatorname{Exp}\left(s_R a_{t,4:6}\right),
  \label{eq:app-action-scale}
\end{equation}
where $s_p$ and $s_R$ scale translation and axis-angle rotation, respectively.
The seventh component opens the gripper when positive and closes it otherwise.
Action scales and gripper targets are listed in \cref{tab:app-parameters}.
\subsection{Actor Observations and Task Encoding}
\label{app:observations}

\begin{figure}[htbp]
  \centering
  \includegraphics[width=0.85\linewidth]{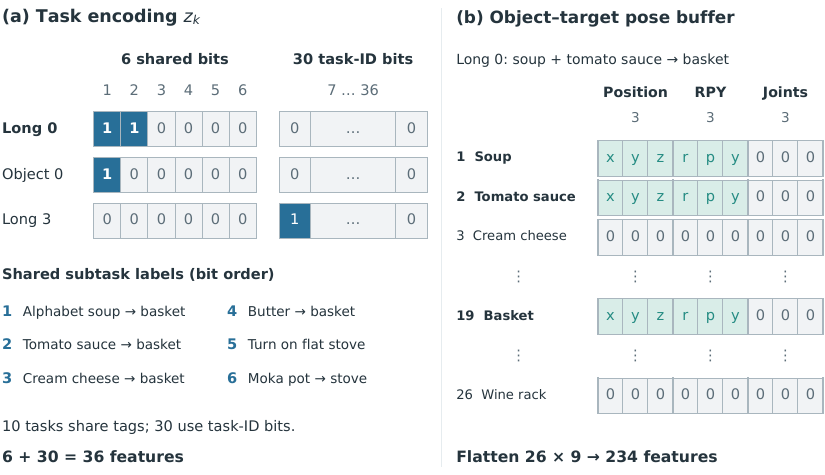}
  \caption{\textbf{Task encoding and object--target pose layout.} (a) Long~0 places alphabet soup and tomato sauce in the basket, activating shared bits 1 and 2. Object~0 activates only bit 1; Long~3 uses a dedicated task-ID bit. (b) The same Long~0 task fills the soup, tomato sauce, and basket slots. Symbols denote live pose coordinates; gray cells are zero, including the joint coordinates of these rigid entities. Row numbers indicate canonical entity slots; ellipses omit intermediate entries.}
  \label{fig:app-observation-encoding}
\end{figure}

\Cref{tab:app-observations} summarizes the actor inputs in \cref{eq:actor-observations}, where square brackets concatenate features.
\Cref{fig:app-observation-encoding} illustrates the task encoding and object--target pose buffer.

\begin{table}[ht]
  \centering
  \caption{\textbf{Benchmark observation groups and input dimensions.} Dimensions use the full 40-task configuration with the default multi-hot encoding. State and visual actors share 66 features of task context and proprioception; their task-state interfaces contribute 234 and 384 features, respectively. Symbols follow \cref{eq:actor-observations}.}
  \label{tab:app-observations}
  \normalsize
  \setlength{\tabcolsep}{4pt}
  \begin{tabularx}{\linewidth}{@{}ll>{\raggedright\arraybackslash}Xlr@{}}
    \toprule
    \textbf{Group} & \textbf{Symbol} & \textbf{Contents} & \textbf{Actor} & \textbf{Dimension} \\
    \midrule
    Task context & $a_{t-1},\ z_k$ & Previous action and task encoding & both & $7+36=43$ \\
    Proprioception & $o_t^{\mathrm{proprio}}$ & End-effector pose (7), gripper joints (2), arm joint positions (7) and velocities (7) & both & $23$ \\
    State task input & $o_t^{\mathrm{object\text{-}target}}$ & Object--target pose buffer: 26 entity slots, 9 features per slot & state & $26\times9=234$ \\
    Visual task input & $h_t^{\mathrm{visual}}$ & Pooled frozen-ViT features from third-person and wrist images & visual & $2\times192=384$ \\
    \midrule
    \multicolumn{4}{@{}l}{State actor total: $43+23+234$} & $\mathbf{300}$ \\
    \multicolumn{4}{@{}l}{Visual actor total: $43+23+384$} & $\mathbf{450}$ \\
    \bottomrule
  \end{tabularx}
\end{table}

\paragraph{Task encoding.}
The default multi-hot vector $z_k\in\{0,1\}^{36}$ combines six shared subtask indicators with 30 dedicated task-ID bits (\cref{fig:app-observation-encoding}a).
Ten tasks activate their applicable shared indicators; each remaining task activates one dedicated bit, ordered lexicographically by its suite-qualified task key.
The task ID $k$ remains fixed throughout the rollout; a canonical one-hot task-ID encoding is also available.

\paragraph{Object--target pose buffer.}
At each step, $o_t^{\mathrm{object\text{-}target}}$ concatenates live object and target states in the robot base frame using the fixed entity slots in \cref{fig:app-observation-encoding}b.
Each slot allocates three coordinates each to position, roll--pitch--yaw orientation, and articulation, with unused joint coordinates and task-inactive slots set to zero.
Focused encoding retains only entities in the active task subset.

\paragraph{Actor input dimensions.}
Both actors share the previous action, task encoding, and 23-dimensional proprioception (end-effector position and quaternion in the robot base frame, gripper finger positions, and arm-joint positions and velocities).
Adding the pose buffer gives 300 state-input features; using the pooled visual feature instead gives 450 (\cref{tab:app-observations}).
Each $224\times224$ camera image yields 196 patch tokens of width 192; concatenating the two views gives a $196\times384$ array, which attention pooling reduces to $h_t^{\mathrm{visual}}\in\mathbb{R}^{384}$ (\cref{app:visual-details}).
The DGPO reference stack's privileged critic inputs are specified separately in \cref{sec:shared-protocol}.
\FloatBarrier

\subsection{Joint-Limit Penalties}
\label{app:task-reward-details}

In \cref{eq:task-reward}, the joint-limit penalties use position bounds $[q_j^{\min},q_j^{\max}]$ and velocity limit $\dot q_j^{\max}$:
\begin{equation}
\begin{aligned}
  d_{j,t}^{\mathrm{pos}}
  &={}[q_j^{\min}-q_{j,t}]_+ + [q_{j,t}-q_j^{\max}]_+, \\
  d_{j,t}^{\mathrm{vel}}
  &=\operatorname{clip}\!\left(
      |\dot q_{j,t}|-0.95\dot q_j^{\max},0,1
    \right).
\end{aligned}
\label{eq:task-safety}
\end{equation}
Here $[x]_+=\max(x,0)$, and the safety sums span all robot joints.
All reward coefficients are listed in \cref{tab:app-parameters}.
\section{DGPO Framework and Reference Learners}
\label{app:dgpo-reference}

This section details the asymmetric critic inputs and visual features, IW-ABC action targets and weighting schedules, and reference learner implementations.

\subsection{Asymmetric Critic Inputs and Visual Features}
\label{app:visual-details}

Actor and critic inputs follow \cref{eq:actor-observations,sec:shared-protocol}; network dimensions are listed in \cref{tab:app-parameters}.
For environment $e$ paired with demonstration $\xi(e)$ of length $T_{\xi(e)}$, let $r_e$ be its zero-based reset cursor and $t$ the number of control steps since reset.
The current demonstration cursor and its normalized critic input are
\begin{equation}
  m_e(t)=\min\!\left(r_e+t,\,T_{\xi(e)}-1\right),
  \qquad
  \phi_{e,t}=\frac{m_e(t)}{\max\!\left(T_{\xi(e)}-1,1\right)}\in[0,1],
  \label{eq:normalized-demo-cursor}
\end{equation}
where $r_e$ and $T_{\xi(e)}$ are measured in stored control steps; $\phi_{e,t}$ is written as $\phi_t$ when the environment index is omitted.

\noindent\textbf{Visual feature aggregation.}
A frozen Theia-style Tiny ViT~\citep{shang2024theia} extracts $P$ patch tokens of dimension $D_v$ from each third-person and wrist image.
The two views are concatenated along the feature dimension into $Z_t\in\mathbb{R}^{P\times 2D_v}$ and reduced by the actor's trainable single-query attention pool:

\begin{equation}
  h_t^{\mathrm{visual}}
  =
  \sum_{p=1}^{P}
  \operatorname{softmax}_p
  \left(\frac{q_\pi^\top Z_{t,p}}{\sqrt{2D_v}}\right)Z_{t,p},
  \qquad q_\pi\in\mathbb{R}^{2D_v}.
  \label{eq:app-attn-pool}
\end{equation}
Here $Z_{t,p}$ is the concatenated token at patch $p$, and softmax normalizes attention over patches.
The pooled vector $h_t^{\mathrm{visual}}\in\mathbb{R}^{2D_v}$ supplies the actor input in \cref{eq:actor-observations}.
The critic applies the same pooling operation with its own query $q_V$, producing $h_t^{V,\mathrm{visual}}$ from the same frozen tokens.
The resulting visual critic input is given in \cref{sec:shared-protocol}.
The Tiny ViT image encoder remains frozen; PPO trains the separate actor and critic attention pools and MLPs.

\subsection{Time-Indexed Action Targets}
\label{app:matched-action}
ABC uses direct time indexing to construct lightweight action targets for massively parallel rollouts, avoiding online state matching, phase estimation, or optimal-transport alignment.
Ordinary training episodes restore the stored initial simulator state $s^D_{\xi(e),0}$ with $r_e=0$; RFCL's reset curriculum is specified in \cref{app:reference-learners}.
Using the cursor $m_e(t)$ from \cref{eq:normalized-demo-cursor}, the ABC action target is
\begin{equation}
  a_{e,t}^{\star}=a^{\mathrm{demo}}_{\xi(e),m_e(t)}.
  \label{eq:time-indexed-action}
\end{equation}
Here $a^{\mathrm{demo}}_{\xi,m}$ is the stored action at index $m$ of trajectory $\xi$; $a_{e,t}^{\star}$ is the environment-specific form of $a_t^{\star}$ in \cref{eq:abc}.
The cursor advances by one index per control step until reaching $T_{\xi(e)}-1$, triggering demonstration-horizon truncation and an environment reset.
This lightweight action target may become misaligned when the policy drifts from the paired demonstration.
Demonstration-initial-state resets establish initial alignment, while the shared tracking reward encourages the rollout to remain close to the reference, supporting the target's use as a soft imitation guide.
The coefficient in \cref{eq:beta-base} further reduces imitation pressure as task success increases and anneals it over training, allowing the policy to increasingly depart from the demonstrated actions.

\subsection{IW-ABC Minibatch Weighting and Schedules}
\label{app:dgpo-training}

\begin{figure}[!htbp]
  \centering
  \includegraphics[width=0.9\linewidth]{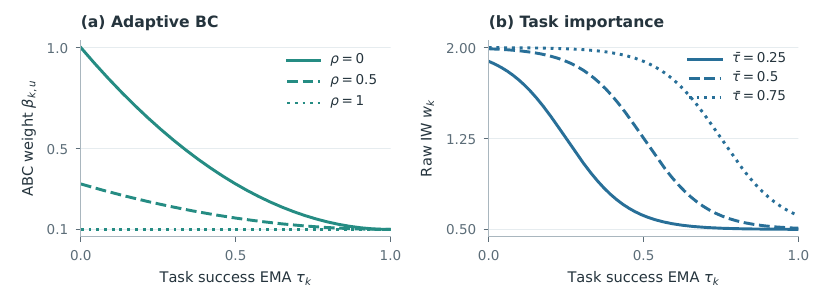}
  \caption{\textbf{Online success controls the two adaptive framework weights.}
  ABC pressure (left; $\beta_{k,u}{\in}[\BCFloor,\BCCeiling]$, $\kappa{=}\BCKappa$) decreases with absolute task success and time, with $\rho_u=1$ after \BCHorizonSteps{} vectorized control steps.
  Raw task importance (right; $w_k{\in}[\IWLower,\IWUpper]$, $s_{\mathrm{IW}}{=}\IWSlope$) compares each task with the mean over $\mathcal{K}_{\mathrm{init}}$ and is minibatch-normalized. Both use the shared default in \cref{tab:app-parameters}.}
  \label{fig:weight-schedules}
\end{figure}

For minibatch $\mathcal{B}$, sample $i$ belongs to task $k(i)$ and uses $\tilde w_k=w_k/(|\mathcal{B}|^{-1}\sum_{i\in\mathcal{B}}w_{k(i)})$ in \cref{eq:iw-abc}.
This normalization gives unit mean weight across samples, preserving the average loss scale while changing relative task contributions.
The weighted losses are reduced in one shared backward pass, without computing separate per-task gradients.
Warm-up settings are listed in \cref{tab:app-parameters}; \cref{fig:weight-schedules} visualizes the schedules defined in \cref{eq:beta-base,eq:iw}.

\subsection{FAMO-ABC Implementation}
\label{app:famo-abc}

FAMO-ABC adapts the critic-balancing rule from FAMO's original actor--critic implementation~\citep{liu2023famo} to the shared PPO backbone.
It retains the ABC targets and coefficient schedule from \cref{app:matched-action,app:dgpo-training}, while the PPO policy surrogate and entropy use uniform sample weights.
FAMO supplies the critic objective in place of the IW-weighted value loss.

For each PPO minibatch, let $L_k$ be the mean clipped value-regression error for task $k$ and $D_k=L_k+\epsilon$, with $\epsilon=10^{-8}$.
Task loss sums and sample counts are aggregated across GPUs; the active task set $\mathcal{K}_{\mathcal{B}}$ contains tasks represented in the global minibatch.
With trainable logits $\mathbf{q}$ and $\omega_k^{\mathrm{FAMO}}=\operatorname{softmax}(\mathbf{q}_{\mathcal{K}_{\mathcal{B}}})_k$, the value objective is
\begin{equation}
  c=\operatorname{sg}\!\left(\sum_{k\in\mathcal{K}_{\mathcal{B}}}\frac{\omega_k^{\mathrm{FAMO}}}{D_k}\right),
  \qquad
  \mathcal{L}_{V}^{\mathrm{FAMO}}
  =c\sum_{k\in\mathcal{K}_{\mathcal{B}}}\operatorname{sg}(\omega_k^{\mathrm{FAMO}})\log D_k,
  \label{eq:app-famo-value}
\end{equation}
where $\operatorname{sg}$ stops gradients.
This multiplicative scaling follows the FAMO RL implementation and gives effective critic coefficients $c\omega_k^{\mathrm{FAMO}}/D_k$.

After the network update, an additional critic forward pass recomputes $D_k$ on the same observations with fixed returns and old value predictions.
The logits receive the progress-based gradient
\begin{equation}
  \delta_k=\log D_k^{\mathrm{before}}-\log D_k^{\mathrm{after}},
  \qquad
  g_k=\omega_k^{\mathrm{FAMO}}\!\left(\delta_k-\sum_{j\in\mathcal{K}_{\mathcal{B}}}\omega_j^{\mathrm{FAMO}}\delta_j\right).
  \label{eq:app-famo-update}
\end{equation}
We initialize $\mathbf{q}=\mathbf{0}$ and apply Adam to $\mathbf{q}$ using $\mathbf{g}$, learning rate $0.025$, and weight decay $0.01$ after every PPO minibatch.
Tasks absent from the global minibatch are excluded from the softmax and progress calculation.
The shared policy and critic retain one network backward pass per minibatch; FAMO adds the critic forward pass and the task-logit update.

\subsection{Reference Learner Specifications}
\label{app:reference-learners}

The shared setup and demonstration-interface taxonomy are defined in \cref{sec:shared-protocol,sec:demonstration-interfaces}.
We give only the additional objectives, initialization rules, and reset curriculum below; hyperparameters are collected in \cref{tab:app-parameters}.

\noindent\textbf{Offline demonstration dataset preparation.}
Before training, demonstrations are preprocessed into a shared dataset $\mathcal{D}=\{(s_t^D,o_t^D,a_t^D)\}$, where $s_t^D$ is the stored simulator state and $o_t^D$ is the actor observation including task encoding.
Trajectory and timestep indices are retained: BC pretraining and DAPG use the observation--action pairs, while RFCL uses the stored states $s_{\xi,j}^D$ for curriculum resets.

\noindent\textbf{Offline BC and transfer to PPO.}
Offline BC trains an architecture-matched actor on the full dataset $\mathcal{D}$ using mean-squared action error:

\begin{equation}
  \mathcal{L}_{\mathrm{BC}}(\theta)
  =\frac{1}{|\mathcal{D}|}
  \sum_{(s_t^D,o_t^D,a_t^D)\in\mathcal{D}}
  \left\|\mu_\theta(o_t^D)-a_t^D\right\|_2^2.
  \label{eq:app-bc}
\end{equation}
Here $(o_t^D,a_t^D)$ is a recorded pair from the dataset.
BC$\rightarrow$PPO transfers only the pretrained actor weights, initializes a fresh critic and optimizer, and removes the BC loss when online training begins.
Its online budget matches the other learners; offline pretraining is additional.
\noindent\textbf{DAPG regularization.}
DAPG~\citep{rajeswaran2018dapg} augments policy gradient with a demonstration log-likelihood term evaluated on pairs from the same preprocessed dataset $\mathcal{D}$:
\begin{equation}
  \mathcal{L}^{\mathrm{DAPG}}_i
  =-\lambda_0\lambda_1^i
  \mathbb{E}_{(s^D,o^D,a^D)\sim\mathcal{D}}\!\left[
    \log\pi_\theta(a^D\mid o^D)
  \right],
  \label{eq:app-mtdapg}
\end{equation}
Here $i$ is the PPO-update index, $\lambda_0$ is the initial coefficient, and $\lambda_1$ is its per-update decay factor; task encoding is included in $o^D$.

\noindent\textbf{PPO-based RFCL curriculum.}
Our adaptation of RFCL~\citep{tao2024rfcl} maintains an independent reverse-reset curriculum per task, with on-policy PPO updates and no offline demonstration replay or auxiliary imitation loss.
Unlike the original forward-expansion phase, its second phase samples uniformly from all demonstration initial states of the task.
For task $k$, let $\xi_1,\ldots,\xi_{N_k}$ index its demonstration trajectories.
During the reverse phase, trajectory $\xi_i$ of length $T_{\xi_i}$ initializes its frontier cursor to $c_i=\lfloor p_{\mathrm{init}}T_{\xi_i}\rfloor$.
An environment $e$ paired with $\xi(e)=\xi_i$ samples its reset cursor and stored state as
\begin{equation}
  \begin{aligned}
    r_e&=\min(c_i+\Delta,T_{\xi_i}-1),\qquad s_0=s^D_{\xi_i,r_e},\\
    \Delta&=\min(X,4),\quad X\sim\operatorname{Geom}(p_\Delta),\\
    c_i&\leftarrow\max(c_i-\delta,0)\quad\text{after frontier mastery}.
  \end{aligned}
  \label{eq:app-rfcl-reverse}
\end{equation}
Here $s^D_{\xi_i,j}$ is the stored state at step $j$ of trajectory $\xi_i$, $r_e$ is the reset cursor in \cref{eq:normalized-demo-cursor}, and $X$ is a nonnegative geometric offset.
Frontier mastery requires $m$ consecutive successful frontier rollouts, after which the cursor moves backward by $\delta$.
Once all active cursors for task $k$ reach zero, resets switch to its full set of $N_k$ demonstration starts:
\begin{equation}
  i\sim\operatorname{Uniform}\{1,\ldots,N_k\},
  \qquad
  \xi(e)=\xi_i,\quad r_e=0,\quad s_0=s^D_{\xi_i,0},
  \label{eq:app-rfcl-demo-start}
\end{equation}
Initialization and mastery settings are listed in \cref{tab:app-parameters}.

\section{Reproducibility Configuration}
\label{app:parameter-values}

\Cref{tab:app-parameters} groups the benchmark settings, reference-policy architectures, optimization parameters, and learner-specific schedules by component.
The tracking weights $w_{\mathrm{ee,p}},w_{\mathrm{ee,r}},\ldots$ instantiate the component weights $w_m$ in \cref{eq:demo-tracking-reward}.

\begin{center}
\normalsize
\setlength{\tabcolsep}{3pt}
\begin{longtable}{@{}>{\raggedright\arraybackslash}p{0.22\linewidth}>{\raggedright\arraybackslash}p{0.48\linewidth}>{\raggedright\arraybackslash}p{0.25\linewidth}@{}}
\caption{\textbf{Reproducibility configuration by component.} Settings for the benchmark and reported reference learners, including schedule units.}
\label{tab:app-parameters}\\
\toprule
\textbf{Symbol / setting} & \textbf{Description} & \textbf{Value} \\
\midrule
\endfirsthead
\multicolumn{3}{@{}l}{\tablename~\thetable\ (continued)} \\
\toprule
\textbf{Symbol / setting} & \textbf{Description} & \textbf{Value} \\
\midrule
\endhead
\multicolumn{3}{@{}l}{\emph{Task set and demonstrations}} \\*
$K$ & Jointly trained tasks & $40$ (ten per suite) \\*
$p(T)$ & Task allocation & Uniform \\*
$N_k$ & Demonstration trajectories per task & $50$ \\*
\midrule
\multicolumn{3}{@{}l}{\emph{Benchmark simulation}} \\*
$\Delta t_{\mathrm{sim}}$ & Physics time step & $1/60$ s \\*
$d$ & Action decimation & $3$ \\*
$\Delta t_{\mathrm{ctrl}}$ & Control time step & $0.05$ s (20 Hz) \\*
$H$ & Maximum episode horizon in control steps & $520$ \\*
$T_{\max}=H\Delta t_{\mathrm{ctrl}}$ & Maximum episode duration & $26$ s \\*
$\eta_v$ & Joint velocity limit multiplier & $1.5$ \\
\midrule
\multicolumn{3}{@{}l}{\emph{Action and controller}} \\*
$\dim(a_t)$ & Policy action dimension & $7$ \\*
$s_p$ & Relative translation action scale & $1/20$ \\*
$s_R$ & Relative rotation action scale & $1/2$ \\*
$\theta^{\mathrm{grip}}_{\mathrm{open}}$ & Open gripper target & $0.04$ m \\*
$\theta^{\mathrm{grip}}_{\mathrm{close}}$ & Closed gripper target & $0.0$ m \\
\midrule
\multicolumn{3}{@{}l}{\emph{Benchmark task reward and safety}} \\*
$w_{\mathrm{succ}}$ & Sparse success bonus & $1.0$ \\*
$w_{\Delta a}$ & Action smoothness penalty & $5{\times}10^{-4}$ \\*
$w_a$ & Action norm penalty & $5{\times}10^{-4}$ \\*
$w_{\dot q}$ & Joint velocity penalty & $10^{-3}$ \\*
$w_{\mathrm{pos}}^{\mathrm{lim}}$ & Joint position soft-limit penalty & $1.0$ \\*
$w_{\mathrm{vel}}^{\mathrm{lim}}$ & Joint velocity soft-limit penalty & $0.5$ \\
\midrule
\multicolumn{3}{@{}l}{\emph{DGPO demonstration-tracking reward}} \\*
$\sigma_{r,m}$ & Component-specific tracking kernel bandwidth & $0.1$ \\*
$w_{\mathrm{ee,p}}$ & Dense EE position tracking weight & $0.5$ \\*
$w_{\mathrm{ee,r}}$ & Dense EE rotation tracking weight & $1.0$ \\*
$w_{\mathrm{grip}}$ & Dense gripper tracking weight & $0.4$ \\*
$w_{\mathrm{joint,p}}$ & Dense arm-joint position tracking weight & $0.5$ \\*
$w_{\mathrm{joint,v}}$ & Dense arm-joint velocity tracking weight & $0.1$ \\*
$w_{\mathrm{obj,p}}$ & Dense object position tracking weight & $0.10$ \\*
$w_{\mathrm{art}}$ & Dense articulation tracking weight & $0.05$ \\*
$w_{\mathrm{obj,r}}$ & Dense object rotation tracking weight & $0.05$ \\
\midrule
\multicolumn{3}{@{}l}{\emph{Shared state/visual IW-ABC default schedule}} \\*
$u$ & Schedule counter accumulated across rollouts & Vectorized control steps \\*
$s_{\mathrm{IW}}$ & Task-importance sigmoid slope & $\IWSlope$ \\*
$w_{\max}$ & Maximum importance weight & $\IWUpper$ \\*
$w_{\min}$ & Minimum importance weight & $\IWLower$ \\*
$\alpha_{\mathrm{ema}}$ & Per-task success EMA rate & $\TaskEMAAlpha$ \\*
$N_{\mathrm{reset}}$ & Completed episodes per task required for inclusion in $\mathcal{K}_{\mathrm{init}}$ & $20$ \\*
$U_{\mathrm{IW}}$ & Task-importance warmup (vectorized control steps) & $\IWStartSteps$ \\*
$\beta_{\max}$ & Maximum ABC weight & $\BCCeiling$ \\*
$\beta_{\min}$ & Minimum ABC weight & $\BCFloor$ \\*
$\kappa$ & Adaptive-BC beta-kernel concentration & $\BCKappa$ \\*
$U_{\mathrm{ann}}$ & ABC horizon (vectorized control steps) & $\BCHorizonSteps$ \\*
$U_{\mathrm{ann}}/T_{\mathrm{rollout}}$ & Equivalent ABC horizon in PPO iterations & $\BCHorizonIterations$ \\*
$c_{\mathrm{BC}}$ & Base ABC coefficient & $\BCBase$ \\
\midrule
\multicolumn{3}{@{}l}{\emph{Shared reference-policy architecture}} \\*
$\dim(o_t^{\mathrm{state}})$ & State actor input dimension & $300$ \\*
$h_\pi,h_V$ & Actor and critic MLP hidden widths & $[512,256,128]$ \\*
$\mathrm{act}$ & MLP activation & ELU \\*
History & Temporal context & None (feed-forward) \\*
$\sigma_0$ & Initial policy standard deviation (log-parameterized Gaussian) & $0.8$  \\*
$\mathrm{norm}$ & Observation normalization & running mean/std \\
\midrule
\multicolumn{3}{@{}l}{\emph{Shared PPO reference stack}} \\*
Training horizon & Online PPO iterations & $30{,}000$ \\*
$\mathrm{opt}$ & Optimizer & Adam, $\beta_1{=}0.9$, $\beta_2{=}0.999$ \\*
$\gamma$ & Discount factor & $0.99$ \\*
$\varepsilon$ & PPO policy and value clip range & $0.15$  \\*
$c_V$ & Value-loss coefficient & $1.0$ \\*
$c_H$ & Entropy bonus coefficient & $0.005$  \\*
$\lambda_{\mathrm{GAE}}$ & GAE trace decay & $0.95$ \\*
$T_{\mathrm{rollout}}$ & Vectorized control steps per PPO iteration & $\PPORolloutSteps$ \\*
$N_{\mathrm{ep}}$ & PPO epochs per rollout & $5$ \\*
$N_{\mathrm{mb}}$ & Minibatches per epoch & $4$ \\*
$\eta_\theta$ & Policy and value learning rate & $2{\times}10^{-4}$ \\*
$\mathrm{KL}^\star$ & Adaptive learning rate KL target & $0.005$ \\*
$g_{\max}$ & Global gradient norm clip & $1.0$ \\
\midrule
\multicolumn{3}{@{}l}{\emph{Visual reference configuration}} \\*
Backbone & Tiny ViT training and execution mode & Frozen; inference mode \\*
Cameras & RGB observation views & Third-person and wrist \\*
$p_{\mathrm{vit}}$ & ViT patch size & $16$ \\*
$D_v$ & ViT hidden dimension & $192$ \\*
$L_v$ & ViT transformer depth & $12$ \\*
$H_{\mathrm{img}}, W_{\mathrm{img}}$ & Camera image resolution & $224, 224$ \\*
$P$ & Patch tokens per camera & $196$ \\*
\midrule
\multicolumn{3}{@{}l}{\emph{Offline BC and BC$\rightarrow$PPO pretraining}} \\*
$|\mathcal{D}|$ & Demonstration state--observation--action records & $336{,}575$ \\*
Training horizon & Offline optimizer steps & $30{,}000$ \\*
Batch size & Transitions per optimizer step & $4{,}096$ \\*
Learning rate & Offline actor learning rate & $10^{-4}$ \\*
Normalization & Actor observations & Running mean/std \\
\midrule
\multicolumn{3}{@{}l}{\emph{DAPG offline regularization}} \\*
$\lambda_0$ & DAPG initial coefficient & $0.1$ \\
$\lambda_1$ & DAPG iteration decay & $0.995$ \\
\midrule
\multicolumn{3}{@{}l}{\emph{RFCL reset curriculum}} \\*
$m$ & RFCL frontier success window & $3$ rollouts \\*
$\delta$ & RFCL cursor decrement & $8$ steps \\*
$p_\Delta$ & RFCL geometric offset parameter & $0.5$ \\*
$p_{\mathrm{init}}$ & RFCL Phase~1 cursor init fraction & $0.85$ \\
\bottomrule
\end{longtable}
\end{center}

\section{Supplementary IW-ABC Analyses}
\label{app:supp-experiments}

We examine where to apply task importance weighting in the actor and critic, and how IW and ABC weights evolve across tasks during training.

\subsection{Task Weighting in the Actor and Critic}
\label{app:sim-experiments}

The common task allocation, policy class, demonstration set, online PPO horizon, and training seeds are specified in \cref{sec:setup}; numerical optimization and rollout settings are consolidated in \cref{tab:app-parameters}.
The ablation below tests where task IW is applied within the selected IW-ABC reference configuration while retaining the same learner family.

\paragraph{Where task IW is applied.}
Within IW-ABC, \cref{tab:app-iw-placement} compares policy-side weighting with the default that also weights value regression.
Using the minibatch notation from \cref{app:dgpo-training} and coefficients from \cref{eq:iw-abc}, 
let $\ell_i^{\pi}$ and $\ell_i^V$ denote the per-sample negative clipped PPO surrogate loss and clipped value-regression loss at minibatch index $i$, respectively; $\mathcal{H}_i$ is the policy entropy.
The ABC demonstration term in \cref{eq:abc} is unchanged and omitted below; the two PPO variants optimize
\begin{align}
  \mathcal{L}_{\mathrm{actor+critic}}
  &= \frac{1}{|\mathcal{B}|}\sum_{i\in\mathcal{B}}
  \tilde w_{k(i)}\!\left(\ell_i^{\pi}+c_V\ell_i^V-c_H\mathcal{H}_i\right), \\
  \mathcal{L}_{\mathrm{actor\ only}}
  &= \frac{1}{|\mathcal{B}|}\sum_{i\in\mathcal{B}}
  \left[\tilde w_{k(i)}\!\left(\ell_i^{\pi}-c_H\mathcal{H}_i\right)+c_V\ell_i^V\right].
  \label{eq:iw-placement}
\end{align}
Thus entropy follows the actor weight in both variants; all other settings are identical.
Weighting both actor and critic improves Mean SR by 11.9 percentage points and Long SR by 20.0 points over actor-only weighting, while increasing task coverage from 31 to 35 out of 40 (\cref{tab:app-iw-placement}).
These gains support applying task IW to value regression as well as policy optimization, which we adopt as the default.

\begin{table}[ht]
  \centering
  \caption{\textbf{Task-IW placement.}
  The reported ablation uses IW-ABC with a PPO backbone. ``Actor'' includes policy and entropy; ``critic'' denotes value regression.}
  \label{tab:app-iw-placement}
  \normalsize
  \setlength{\tabcolsep}{3.5pt}
  \begin{tabular}{lccrrr}
    \toprule
    \textbf{IW placement} & \textbf{Actor} & \textbf{Critic} & \makecell{\textbf{Mean}\\\textbf{SR (\%)}} & \makecell{\textbf{Long}\\\textbf{SR (\%)}} & \makecell{\textbf{Task}\\\textbf{coverage}} \\
    \midrule
    Actor only & \cmark & \xmark & 78.2$\pm$3.1 & 50.0$\pm$0.0 & 31/40 \\
    \textbf{Actor + critic (default)} & \cmark & \cmark & \textbf{90.1$\pm$3.8} & \textbf{70.0$\pm$10.0} & \textbf{35/40} \\
    \bottomrule
  \end{tabular}
\end{table}

\FloatBarrier
\subsection{Task-Level IW and ABC Dynamics}
\label{app:weight-dynamics}

\begin{figure}[!htbp]
  \centering
  \includegraphics[width=\linewidth]{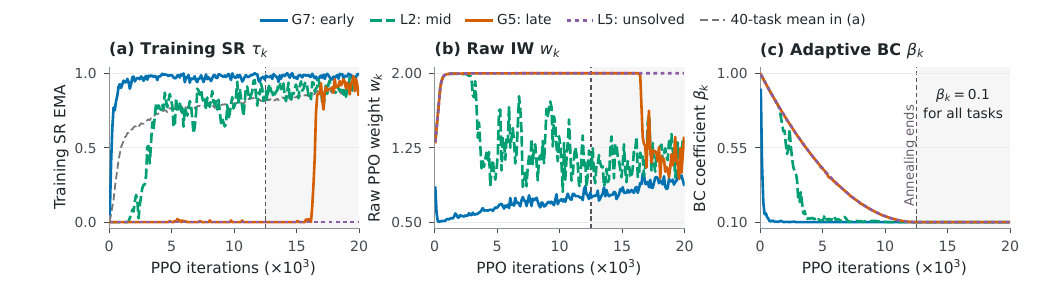}
  \caption{\textbf{Illustrative task trajectories for Vis IW-ABC.}
  Training success EMA $\tau_k$ (a), raw IW $w_k$ (b), and BC coefficient $\beta_{k,u}$ (c) for G7, L2, G5, and L5 from the same single training run as \cref{fig:weight-dynamics}.
  Gray in (a) denotes the 40-task mean.
  Vertical dashed lines mark the end of ABC annealing at $\BCHorizonSteps$ vectorized control steps, after which every task has $\beta_{k,u}=\BCFloor$.}
  \label{fig:app-weight-trajectories}
\end{figure}

\Cref{fig:app-weight-trajectories} complements \cref{fig:weight-dynamics} with four tasks selected to illustrate different learning speeds in the same Vis IW-ABC run.
G7 learns early, reducing both its BC coefficient and its relative PPO weight, while L2 improves during annealing and progressively relinquishes demonstration guidance.
G5 retains high IW after BC reaches its residual floor; success rises around 17k iterations, followed by lower IW as the task approaches the multi-task mean.
L5 remains unsolved within the observed training budget and retains high IW, maintaining its PPO priority.
Together, these examples show how ABC relaxes imitation with task progress and training time, while IW continues emphasizing tasks that lag behind.

\section{Real-World Deployment Details}
\label{app:real-world}
\raggedbottom

\paragraph{Deployment pipeline.}
FoundationPose~\citep{wen2024foundationpose} estimates object poses from RGB-D observations using object meshes corresponding to the simulation USD assets.
Camera calibration and object-frame alignment express these estimates in the robot base frame.
The planar target pad uses color segmentation and depth back-projection instead of FoundationPose.
The frozen actor produces joint-position targets and gripper commands in a closed feedback loop.

\paragraph{Piper observation layout.}
The Piper policy uses a $42$-dimensional object--target buffer with six slots, each storing a 3D position and a four-component quaternion; task-inactive slots are zero.
Together with task identity (4), previous action (7), joint positions (8) and joint velocities (8), this yields a $69$-dimensional actor input matching the Piper training interface.

\flushbottom

\end{document}